\documentclass{article}
\pdfoutput=1
\PassOptionsToPackage{numbers}{natbib}
\usepackage[final]{nips_2016}

\usepackage[T1]{fontenc}    
\usepackage{url}            
\usepackage{booktabs}       
\usepackage{amsfonts}       
\usepackage{nicefrac}       
\usepackage{microtype}      
\usepackage{algorithm}
\usepackage{amsmath}
\usepackage{graphicx}
\usepackage{algpseudocode}
\usepackage{float}
\usepackage{amsopn}
\usepackage{amsmath}

\usepackage{pifont}
\newcommand{\cmark}{\ding{51}}
\newcommand{\xmark}{\ding{55}}
\usepackage[textsize=tiny]{todonotes}
\DeclareMathOperator{\Softmax}{Softmax}

\title{Learning to Reason with Adaptive Computation}

\author{
    Mark Neumann\thanks{Currently at the Allen Institute for Artificial Intelligence.} \hspace{0.66em} Pontus Stenetorp \hspace{0.66em} Sebastian Riedel \\
    \texttt{mark.neumann.15@ucl.ac.uk} \hspace{0.66em}
    \texttt{\{p.stenetorp,s.riedel\}@cs.ucl.ac.uk} \\
    University College London, London, United Kingdom \\
}
\begin{document}

\maketitle

\begin{abstract}
    Multi-hop inference is necessary for machine learning systems to successfully solve tasks such as Recognising Textual Entailment and Machine Reading.
    In this work, we demonstrate the effectiveness of adaptive computation for learning the number of inference steps required for examples of different complexity and that learning the correct number of inference steps is difficult.
    We introduce the first model involving Adaptive Computation Time which provides a small performance benefit on top of a similar model without an adaptive component as well as enabling considerable insight into the reasoning process of the model.
\end{abstract}

\section{Introduction}

Recognising Textual Entailment (RTE) is the task of determining whether a hypothesis can be inferred from a premise.
We argue that natural language inference requires the combination of inferences and aim to provide a stepping stone towards the development of such a method.
These steps can be compared to backtracking in a logic programming language and by employing an attention mechanism we are able to visualise each inference step, allowing us to interpret the inner workings of our model.

At the centre of our approach is Adaptive Computation Time~\cite{ACT}, which is the first example of a neural network defined using a static computational graph which can execute a variable number of inference steps which is conditional on the input.
However, ACT was originally only applied to a vanilla Recurrent Neural Network, whereas we show that it can be applied to arbitrary computation for a problem not explicitly defined to benefit from ACT.
 
\section{A Motivating Multi-hop Inference Example}

When humans resolve entailment problems, we are adept at not only breaking down large problems into sub-problems, but also re-combining the sub-problems in a structured way. 
Often in simple problems -- such as negation resolution -- only the first step is necessary.
However, in more complicated cases -- such as multiple co-reference resolution, or lexical ambiguity -- it can be necessary to both decompose and then reason.
For instance the contradicting statements:

\begin{itemize}
    \item{\textbf{Premise:} An elderly gentleman stands near a bus stop, using an umbrella for shelter because there is a thunderstorm.}
    \item{\textbf{Hypothesis:} An old man holding a closed umbrella is sheltering from bad weather under a bus stop.}
\end{itemize}

require multi-step, \textit{temporally dependent} reasoning to resolve correctly.
On closer examination, the key to resolution here is the action relating entities and conditions in the scene, leading to an inference chain similar to:

\begin{itemize}
    \item{An old man, an umbrella and a bus stop are present \cmark}
    \item{The weather is bad \cmark}
    \item{He is sheltering from the weather \cmark}
    \item{He is sheltering from the weather using a \textit{closed umbrella} \xmark}
\end{itemize}

where the final true/false statement is built up from first observing facts about the scene and then combining and extending them.
We use this idea of combining distinct low-level inferences and show that Adaptive Computation and multi-step inference can be used to examine how incorporating additional inferences into the inference chain can influence the final classification.
The visualisation of which provides an additional tool for the analysis of deep learning based models.

\section{Approach}

Attention mechanisms for neural networks, first introduced for Machine Translation~\cite{align_and_translate}, have demonstrated state-of-the-art performance for RTE~\cite{reasoning_about_entailment,DA}.
Our model is an extension of the Iterative Alternating Attention model originally employed for Machine Reading,~\cite{iterative_neural_attention_mr} combined with the Decomposable Attention (DA) model previously proposed for RTE.~\cite{DA}
The original motivation behind the Decomposable Attention model was to bypass the bottleneck of generating a single document representation, as this can be prohibitively restrictive when the document grows in size.
Instead, the model incorporates an ``inference''' step, in which the query and the document are iteratively attended over for a fixed number of iterations in order to generate a representation for classification.

Our contribution is to generalise this inference step to run for an adaptive number of steps conditioned on the input using a single step of the Adaptive Computation Time algorithm for Recurrent Neural Networks.
We also show that ACT can be used to execute arbitrary functions by considering it as a (nearly) differentiable implementation of a while loop (Appendix~\ref{ACT}).

\textbf{BOW Attention Encoders:} As in the original DA formulation, the vector representations of words are simply generated by the dot product of the result of a single feedforward network appended to the original word vector. 
The un-normalised alignment weight given hypothesis and premise representations $h_{1}...h_{m}$ and $p_{1}...p_{n}$ are defined as:
\begin{equation}
     e_{ij} = F(h_{i})^{T}F(p_{j})
\end{equation}

Where $F: \mathbb{R}^{d} \rightarrow \mathbb{R}^{d}$ is a feedforward network with ReLU activation functions.
The vector representations of the hypothesis and premise are then generated by taking the softmax over the respective dimensions of the E matrix and concatenating with the original word vector representation:

\begin{equation}
\begin{aligned}
    \beta_{j}& = \sum_{i=1}^{m} \Softmax_{1...m}(e_{ij}) \cdot h_{i} \qquad
    \{\tilde{p}_{j} \}_{i=1}^{n} = [\beta_{j}, p_{j}] \\
    \alpha_{i}& = \sum_{j=1}^{n} \Softmax_{1...n}(e_{ij})\cdot p_{j} \qquad
    \{\tilde{h}_{i} \}_{i=1}^{m} = [\alpha_{i}, h_{i}] \\
\end{aligned}
\end{equation}

\textbf{Alternating Iterative Attention:} Now that we have a representation of both the hypothesis and the premise, we iteratively attend over them.
At inference iteration $t$, we generate an attention representation of the hypothesis:

\begin{equation}
\begin{aligned}
    q_{it}& = \Softmax_{1...m}(\tilde{h}_{i}^{T}(W_{h}s_{t-1} + b_{h})) \\
    q_{t}& = \sum_{i} q_{it} \tilde{h}_{i}
\end{aligned}
\end{equation}

Where $q_{it}$ are the attention weights, $W_{h} \in \mathbb{R}^{d \times s}$ where $s$ is the dimensionality of the \textit{inference GRU state}.
This attention representation of the hypothesis is then used to generate an attention mask over the premise:

\begin{equation}
    d_{it} = \Softmax_{1...n}(\tilde{p}_{i}^{T}(W_{p}[s_{t-1}, q_{t}] + b_{p})) \qquad
    d_{t} = \sum_{i} d_{it} \tilde{p}_{i}
\end{equation}

Where $d_{it}$ are the attention weights, $W_{p} \in \mathbb{R}^{d \times(d + s)}$ where $s$ is the dimensionality of the \textit{inference GRU state} and $[x,y]$ denotes the concatenation of vectors.

\textbf{Gating Mechanism:} Although the attention representations could now be concatenated into an input for the inference GRU, we also make one last addition proposed by \cite{iterative_neural_attention_mr} in order to allow attention representations to be forgotten/not used.

\begin{equation}
\begin{aligned}
    \mathbf{r_{t}} &= G_{p}([s_{t-1},d_{t},q_{t}, d_{t}\cdot q_{t}]) \\
    \mathbf{s_{t}} &= G_{h}([s_{t-1},d_{t},q_{t}, d_{t}\cdot q_{t}])
\end{aligned}
\end{equation}

Where $G_{p}, G_{h}$ are 2 layer, feedforward networks $f: \mathbb{R}^{s + 3d} \rightarrow \mathbb{R}^{d}$.
The generated attention representations are multiplied element-wise with the result of the gating function and concatenated($[r_{t} \cdot d_{t} , s_{t} \cdot q_{t}] \in \mathbb{R}^{2d}$), forming the input at time $t$ to the inference GRU.

In \cite{iterative_neural_attention_mr}, the number of inference GRU steps is a hyperparameter of the model.
We instead learn the number of inference steps to take using a single step of the Adaptive Computation Time algorithm, described in Appendix~\ref{ACT}, where the input into the halting layer is the output $y_{t}$ of the inference GRU.

\section{Results}

\begin{figure}[H]
    \centering
    \centerline{\includegraphics[scale = 0.4]{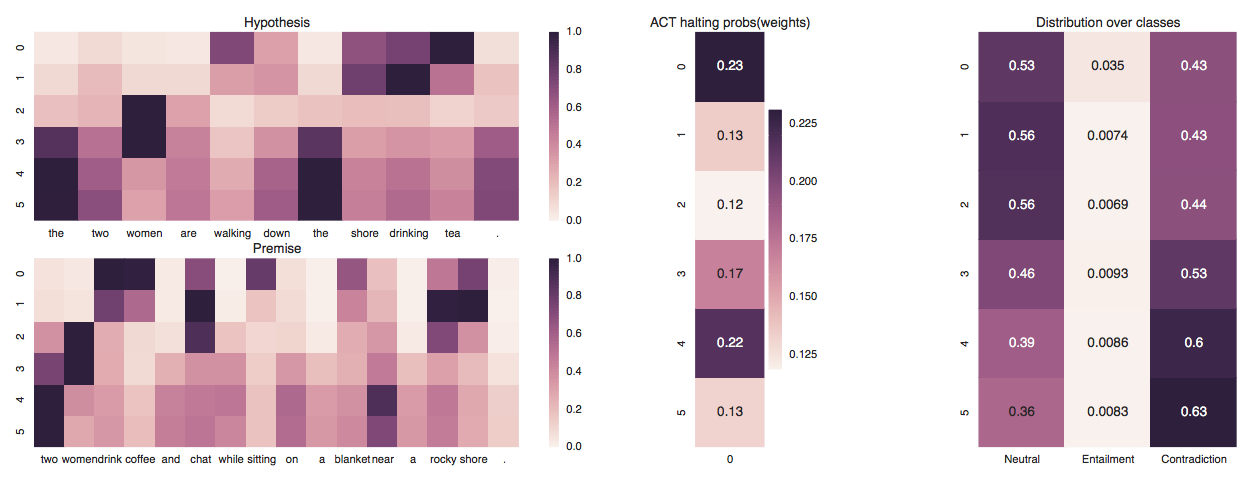}}
    \caption{
        Visualisation of the attention weights produced at each inference step which are used to form the states fed into the inference GRU.
        At each time-step, the hypothesis is attended over using the inference GRU state as the query.
        This representation is then appended to the query and used to generate an attention mask over the Premise.
        On the right, we show the softmax classification if the model had used the inference GRU state at that timestep to classify
        Here, taking multiple steps corrects the classification.
    }
    \label{first_vis}
\end{figure}

In Figure~\ref{first_vis} and Appendix~\ref{extra_figs}, we have provided attention visualisations demonstrating the way the model uses it's access to the premise and hypothesis representations.
We found the following points particularly of note:

\begin{itemize}
    \item{Clear demonstrations of conditional attention, in the sense that the attention weights change over the course of the ACT steps and the model attends over different parts of the sentences.}
    \item{The model very rarely takes a single inference step, even in cases of high regularisation.}
    \item{
        There are several instances of a large ACT weight on the final loop in the inference GRU.
        Demonstrating that when the model has found the key attention representation which provides the necessary insight, it can learn to halt immediately.
    }
\end{itemize}

 Additionally, the Ponder Cost parameter shows context dependent behaviour -- during preliminary experiments using RNN encoders, the optimal Ponder Cost settings were distinct.
 This further reinforces the argument that this parameter is difficult to set correctly. 

Below, we present results on the Stanford Natural Language Inference (SNLI) corpus \cite{SNLI}.

\begin{table}[H]
\centering

\centerline{\begin{tabular}{lcc}
    Model                                                                   & Test Accuracy & Parameters    \\
    \midrule
    Logistic Regression w Lexical features \citep{SNLI}                     & 78.2\%        & n/a           \\
    Baseline LSTM \citep{SNLI}                                              & 77.6\%        & 220k          \\
    1024D "Skip Thought" GRU \citep{order_embeddings}                       & 81.4\%        & 15m           \\
    SPINN-PI Recursive NN \citep{SPINN}                                     & 83.2\%        & 3.7m          \\
    100D LSTM w word-by-word attention \citep{reasoning_about_entailment}   & 83.5\%        & 250k          \\ 
    200D Decomposable Attention \citep{DA}                                  & 86.3\%         & 380k         \\ 
    300D Full tree matching NTI-SLSTM-LSTM \citep{neural_tree_indexers}     & 87.3\%        & 3.3m          \\
    \midrule
    200D Decomposable Attention (our implementation) \citep{DA}             & 83.8\%        & 380k          \\ 
    Adaptive Attention  (ours)                                              & 82.7\%        & 990k          \\
    Adaptive Attention  (2 step)                                            & 76.2\%        & 990k          \\ 
    Adaptive Attention  (4 steps)                                           & 81.7\%        & 990k          \\
    Adaptive Attention  (8 steps)                                           & 81.0\%        & 990k          \\
\end{tabular}}
\caption{
    Table showing relative performance on the test set of the SNLI corpus.
}
\label{test_accs}
\end{table}

\section{Discussion}

It should be noted that we were unable to replicate the performance originally reported for the DA model~\cite{DA} with our implementation \url{https://github.com/DeNeutoy/Decomposable_Attn} despite correspondence with the authors, but continue to seek the reason for this discrepancy. 
However, given our objective was to demonstrate the usefulness of adaptive computation applied to reasoning using attention, we feel this does not detract significantly from the purpose of the paper.
Additionally, our model takes a mean of 5 steps at inference time, meaning it is both more efficient and performant than using a large, fixed number of steps.
The performance plateau demonstrated in Appendix~\ref{more_steps} is most likely an artifact of the way the SNLI corpus is constructed -- in that Amazon Turkers have a financial incentive to provide the minimal hypothesis contradicting, agreeing with, or neutral to the premise.
This construction naturally benefits bag of words approaches, as minimal changes often result in word overlap features being particularly discriminatory.
Additionally our approach involves combining conditional attention masks, which may have less benefit in a situation where there is little difference between the premise and hypothesis.

Overall, we have demonstrated that taking more steps in a multi-hop learning scenario does not always improve performance and that it is desirable to adapt the number of steps to individual data examples. Further exploration into other regularisation methods to tune the ponder cost parameter should be considered, given the model's sensitivity in this regard. The assumption that a single state, rather than combinations or differences of states, is the best method to determine the halting probability should be investigated and will be the subject of future work, as halting promptly demonstrably aids performance. Additionally, we believe that the investigation of attention mechanisms, particularly efficient implementations, such as~\cite{DA,hierarchical_attentive_memory}, when combined with input conditional computation is an important avenue of future research. 

\subsubsection*{Acknowledgments}
This work was supported by a Marie Curie Career Integration Award and an Allen Distinguished Investigator Award. The authors would like to thank Tim Rockt{\"a}schel for helpful feedback and comments.

\bibliographystyle{unsrt}
\bibliography{bibfile}

\begin{thebibliography}{10}

\bibitem{ACT}
Alex Graves.
\newblock Adaptive computation time for recurrent neural networks.
\newblock {\em CoRR}, abs/1603.08983, 2016.

\bibitem{align_and_translate}
Dzmitry Bahdanau, Kyunghyun Cho, and Yoshua Bengio.
\newblock Neural machine translation by jointly learning to align and
  translate.
\newblock {\em CoRR}, abs/1409.0473, 2014.

\bibitem{reasoning_about_entailment}
Tim Rockt{\"a}schel, Edward Grefenstette, Karl~Moritz Hermann, Tom{\'a}s
  Kocisk{\'y}, and Phil Blunsom.
\newblock Reasoning about entailment with neural attention.
\newblock {\em CoRR}, abs/1509.06664, 2015.

\bibitem{DA}
Ankur~P. Parikh, Oscar Tackstrom, Dipanjan Das, and Jakob Uszkoreit.
\newblock A decomposable attention model for natural language inference.
\newblock {\em CoRR}, abs/1606.01933, 2016.

\bibitem{iterative_neural_attention_mr}
Alessandro Sordoni, Phillip Bachman, and Yoshua Bengio.
\newblock Iterative alternating neural attention for machine reading.
\newblock {\em CoRR}, abs/1606.02245, 2016.

\bibitem{SNLI}
Samuel~R. Bowman, Gabor Angeli, Christopher Potts, and Christopher~D. Manning.
\newblock A large annotated corpus for learning natural language inference.
\newblock In {\em EMNLP}, 2015.

\bibitem{order_embeddings}
Ivan Vendrov, Jamie~Ryan Kiros, Sanja Fidler, and Raquel Urtasun.
\newblock Order-embeddings of images and language.
\newblock {\em CoRR}, abs/1511.06361, 2015.

\bibitem{SPINN}
Samuel~R. Bowman, Jon Gauthier, Abhinav Rastogi, Raghav Gupta, Christopher~D.
  Manning, and Christopher Potts.
\newblock A fast unified model for parsing and sentence understanding.
\newblock {\em CoRR}, abs/1603.06021, 2016.

\bibitem{neural_tree_indexers}
Tsendsuren Munkhdalai and Hong Yu.
\newblock Neural tree indexers for text understanding.
\newblock {\em CoRR}, abs/1607.04492, 2016.

\bibitem{hierarchical_attentive_memory}
Marcin Andrychowicz and Karol Kurach.
\newblock Learning efficient algorithms with hierarchical attentive memory.
\newblock {\em CoRR}, abs/1602.03218, 2016.

\bibitem{Adagrad}
John Duchi, Elad Hazan, and Yoram Singer.
\newblock Adaptive subgradient methods for online learning and stochastic
  optimization.
\newblock {\em J. Mach. Learn. Res.}, 12:2121--2159, July 2011.

\bibitem{ADAM}
Diederik~P. Kingma and Jimmy Ba.
\newblock Adam: {A} method for stochastic optimization.
\newblock {\em CoRR}, abs/1412.6980, 2014.

\bibitem{GloVe}
Jeffrey Pennington, Richard Socher, and Christopher~D. Manning.
\newblock Glove: Global vectors for word representation.
\newblock In {\em EMNLP}, 2014.

\end{thebibliography}

\appendix

\section{Does taking more steps improve performance?} \label{more_steps}

Below, we show the accuracy as a function of the fixed inference steps by evaluating the model repeatedly whilst fixing the maximum number of inference steps it can take at test time. From the below graph, it is clear that there is a difference between models using a single inference hop and ones employing more than this. However, performance clearly does not scale linearly with respect to the number of inference hops taken by the model given the similar results for 4-20 hops. We argue that this indifference in performance between inference steps shows that choosing this value in a data dependant way is preferable. Note that the uptick in final performance (when the model may use ACT to determine when to halt) is due to the way ACT combines the inferences using a weighted average over the individual attentions. 

\begin{figure}[H]
    \centering
    \includegraphics[scale = 0.4]{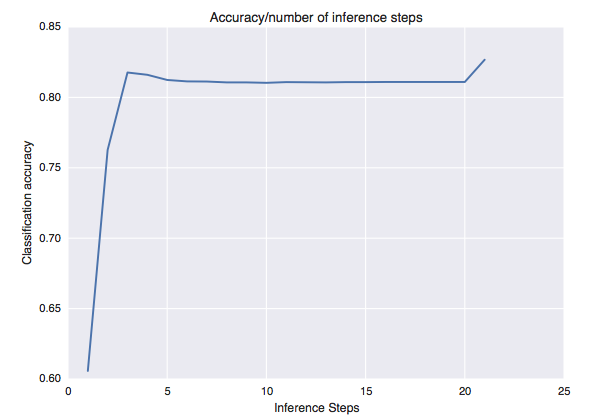}
    \caption{Validation accuracy for the Adaptive Attention model compared to the same model, where the number of steps is fixed at inference time.}
    \label{fig:my_label}
\end{figure}

\section{Additional Diagrams} \label{extra_figs}

\begin{figure}[H]
    \centering
    \centerline{\includegraphics[scale = 0.4]{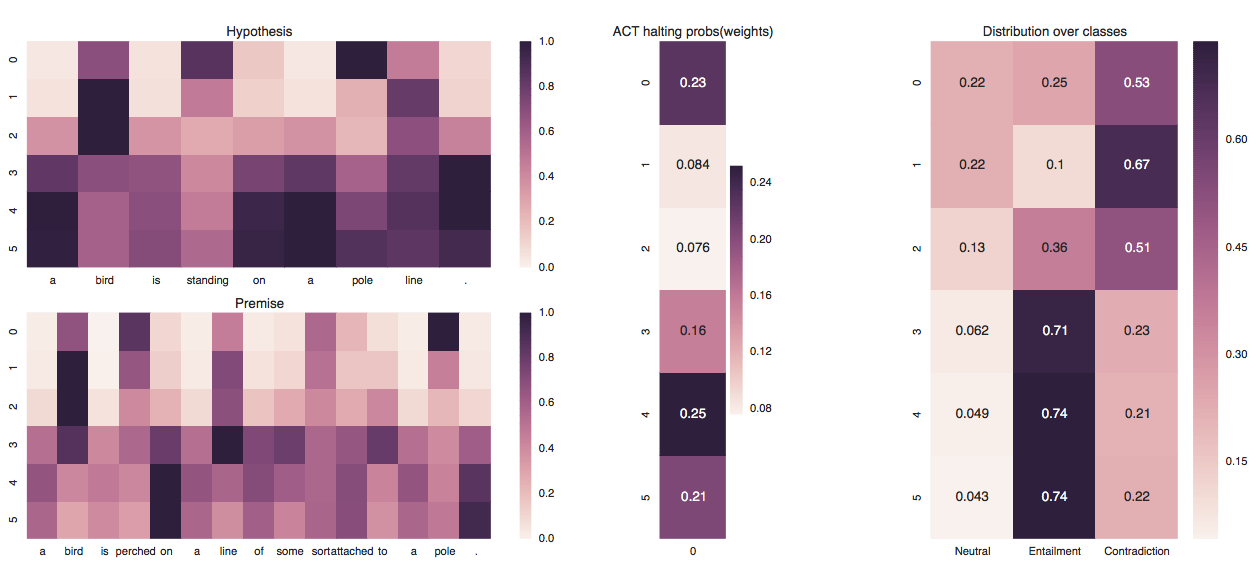}}
    \caption{Here, we observe large final ACT weights once the model has asserted that not only is a bird present (steps 2 to 3), but specifically it is standing \textit{on} the line.}
    \label{fig:birds}
\end{figure}

\begin{figure}[H]
    \centering
    \includegraphics[scale = 0.4]{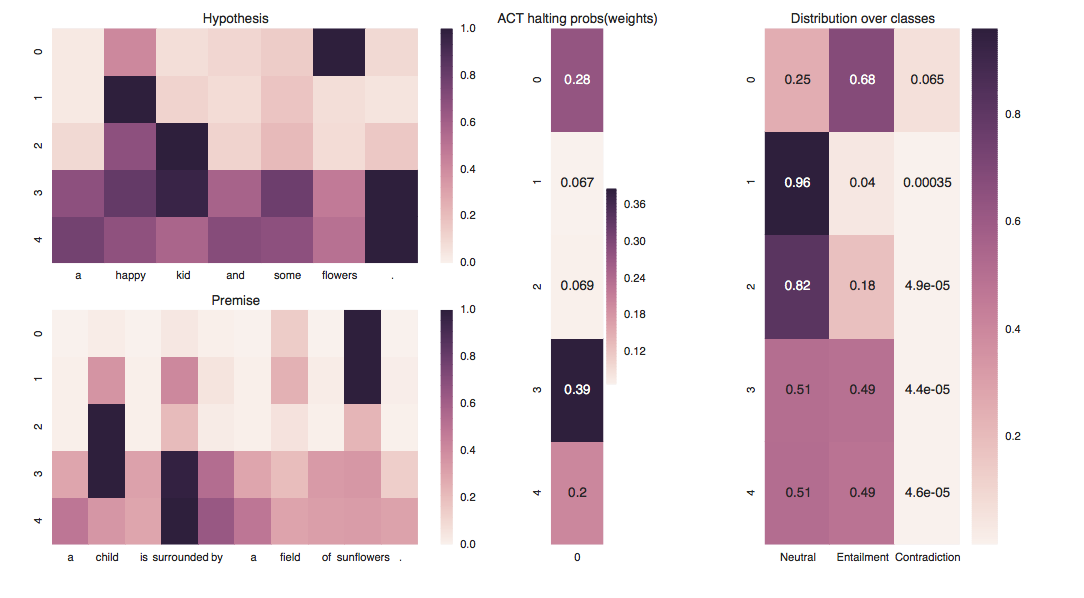}
    \caption{An example of incorrect classification where standard analysis would simply demonstrate a failure - actually, the model is extremely uncertain of whether this falls into the contradiction or neutral categories and only chooses the wrong class by a very small margin. This example additionally provides evidence to support the idea that steps should be adaptive - although this is classified incorrectly, if only a single step were taken here, the example would have been correctly classified.}
    \label{fig:sunflowers}
\end{figure}

\section{Adaptive Computation Time} \label{ACT}

In order to determine the number of steps taken, given a state $s_{t}^{n}$ at time $t$ and inner recurrence step $n$, define:

\begin{equation}
    h_{t}^{n} = \sigma(W_{p}s_{t}^{n} + b_{p}) \qquad
    p_{t}^{n} = \begin{cases}
                    R(t) \qquad n = N(t) \\
                    h_{t}^{n} \qquad otherwise \\
                \end{cases} \\ 
\end{equation}

Where $p_{t}^{n}$ is the halting probability at outer timestep $t$, inner recurrence timestep $n$,  $W_{p}\in \mathbb{R}^{d}, b_{p} \in \mathbb{R}$ are learnt parameters and $\sigma$ is the element-wise sigmoid function.

$N(t)$ is the inner recurrence timestep at which the accumulated halting probabilities $p_{t}^{1...n}$ reach $1- \epsilon$ and $R(t)$ is the remainder at timestep $N(t)$:
\begin{equation}
    N(t) = min \left\{ n : \sum_{i = 1}^{n} p_{t}^{i} \geq 1 - \epsilon \right\} \qquad
    R(t) = 1 - \sum_{i = 1}^{N(t) -1} p_{t}^{i} \\
\end{equation}

Given the above, we define the next cell output and state to be weighted sums of the intermediate states:
\begin{equation}
    s_{t} = \sum_{i=1}^{N(t)}p_{t}^{i} s_{t}^{i} \qquad
    y_{t} = \sum_{i=1}^{N(t)}p_{t}^{i} y_{t}^{i}
\end{equation}

The final addition in \cite{ACT} is the so-called Ponder Cost which is used to regularise the number of computational steps taken.
This is made up of a combination of the remainder function and the iteration counter and added to the original L2 loss function.

\begin{equation}
    \mathcal{P}(x) = \sum_{i=1}^{T} N(t) + R(t)
\end{equation}

Adaptive computation can be seen as a differentiable implementation of a while loop, in that we can perform other operations within the body of a single ACT timestep.
This idea generalises away from the context of ACT within a vanilla recurrent neural network and allowing more complex functionality, such as modules including attention or extraneous inputs.

\begin{algorithm}[H]
\caption{ACT as a while loop}
\label{ACTalgorithm}
\begin{algorithmic}[1]
\Procedure{Inner ACT Step}{}
    \While{$p \leq 1 - \epsilon$}
        \State{$y_{t}^{n}, s_{t}^{n} = InnerCell(x_{n},s_{t}^{n-1})$}
        \State{$p_{t}^{i} = \sigma(W_{p}s_{t}^{n} + b_{p})$}
        \State{$p = p + p_{t}^{i}$}
    \EndWhile{}
    \State{$s_{t} = \sum_{i=1}^{N(t)}p_{t}^{i} s_{t}^{i}$}
    \State{$y_{t} = \sum_{i=1}^{N(t)}p_{t}^{i} y_{t}^{i}$}
\EndProcedure{}
\end{algorithmic}
\end{algorithm}

Note that this algorithmic implementation is for a \textit{single} ACT timestep.

\section{Experimental Details}

In this section, we describe results and insights from running the models described in the Methods section on the Stanford Natural Language Inference Corpus \cite{SNLI}. 

For both the Adaptive Attention model and the Decomposable Attention model, we ran grid searches over the following hyperparameter ranges for 15 epochs over the full training data. 

\begin{table}[H]
\centering
\caption{Hyperparameter ranges for grid searches. \textbf{Bold} and \textit{italicised} text represent the best settings for Adaptive Attention and Decomposable Attention models respectively.}
\label{my-label}
\begin{tabular}{@{}lc@{}}
    Hyperparameter          & Range                                     \\
    \midrule
    Ponder Cost(AA only)    & \textbf{0.001}, 0.0005, 0.0001, 0.00005   \\
    Learning Rate           & AA: \{ \textbf{0.01}, 0.05, 0.001 \} DA: \{0.001, 0.0005,\textit{0.0001} \}  \\
    Dropout                 & 0.1, \textbf{\textit{0.2}}                \\
    Embedding Dim           & \textbf{200}, \textit{256}, 300           \\
    Batch Size              & \textit{8}, 16, \textbf{32}               \\
    Epsilon (AA only)       & \textbf{0.01}, 0.2                        \\
\end{tabular}
\end{table}

Not included in the hyperparameter search were vocab size, fixed to the 40,000 most frequent words (with an OOV token used for words not found in this vocabulary) and the Inference GRU size, fixed to 200. We clipped gradients at an absolute value of 5. All models were trained using the Adagrad optimiser \cite{Adagrad} or the ADAM optimiser \cite{ADAM} respectively for the AA and DA models. Word embeddings are initialised using GloVe pre-trained vector representations \cite{GloVe} (which are normalised for the DA model, following correspondence with the authors) and words without a pre-trained representation are initialised to vectors drawn from a standard Normal distribution $\mathbb{N}(0,0.01)$. Word embeddings are not updated during training, but non-linear embedding projection parameter is trained. 

Following the approach taken by Bowman et al. in the original SNLI paper, we discard any examples with no gold label, leaving us with 549,367, 9,842 and 9,824 for training, validation and testing respectively. After running these hyperparameter searches, the hyperparameter setting with the best accuracy on the validation split is selected and trained until convergence. Only these models are evaluated on the test set.

To generate the comparison between the Adaptive Attention model and the models with fixed numbers of hops, we fix the hyperparameters to the best variant from the grid search and simply halt inference at different fixed numbers of memory hops. 

\end{document}